**Frontiers** | Frontiers in Robotics and AI

# Soft robotics towards sustainable development goals and climate actions

Goffredo Giordano  [1,2]*†, Saravana Prashanth Murali Babu  [3]*†
and Barbara Mazzolai  [1]*

[1]Bioinspired Soft Robotics, Istituto Italiano di Tecnologia (IIT), Genova, Italy, [2]Department of Mechanics
Mathematics and Management, Politecnico di Bari, Bari, Italy, [3]SDU Soft Robotics, SDU Biorobotics, The
Mærsk McKinney Møller Institute, University of Southern Denmark, Odense, Denmark

Soft robotics technology can aid in achieving United Nations' Sustainable Development Goals (SDGs) and the Paris Climate Agreement through development of autonomous, environmentally responsible machines powered by renewable energy. By utilizing soft robotics, we can mitigate the detrimental effects of climate change on human society and the natural world through fostering adaptation, restoration, and remediation. Moreover, the implementation of soft robotics can lead to groundbreaking discoveries in material science, biology, control systems, energy efficiency, and sustainable manufacturing processes. However, to achieve these goals, we need further improvements in understanding biological principles at the basis of embodied and physical intelligence, environment-friendly materials, and energy-saving strategies to design and manufacture self-piloting and field-ready soft robots. This paper provides insights on how soft robotics can address the pressing issue of environmental sustainability. Sustainable manufacturing of soft robots at a large scale, exploring the potential of biodegradable and bioinspired materials, and integrating onboard renewable energy sources to promote autonomy and intelligence are some of the urgent challenges of this field that we discuss in this paper. Specifically, we will present field-ready soft robots that address targeted productive applications in urban farming, healthcare, land and ocean preservation, disaster remediation, and clean and affordable energy, thus supporting some of the SDGs. By embracing soft robotics as a solution, we can concretely support economic growth and sustainable industry, drive solutions for environment protection and clean energy, and improve overall health and well-being.

## 1 Introduction

The advancement of technology has a profound and far-reaching impact on society, currently penetrating all areas of life. However, this advancement negatively affects our ecosystems with growing demands on energy, contributions to greenhouse gas (GHG) emissions, deforestation, and environmental pollution (Arias et al., 2021; Barnard et al., 2021; Armstrong McKay et al., 2022; Kemp et al., 2022; Williams et al., 2022). Mitigating these adverse effects is among the grand challenges of our times and provides a strong motivation to push the research frontier on sustainable materials and robotics to fulfill some of the United Nations' (UNs') Sustainable Development Goals (SDGs) and the Paris Climate Agreement (CPA) (United Nations Environment Programme (12/1), 2016; United Nations' Sustainable Development Goals, 2023). Additionally, to the scientific and technological improvements, achieving SDGs requires a concerted effort from different stakeholders, including governments, researchers, businesses, and civil society organizations. To date, about 3.3–3.6 billion people live in a situation of high risk from climate change (Pörtner et al., 2022). The Intergovernmental Panel on Climate Change formulated in its Sixth Assessment in 2022 that in terrestrial ecosystems, 3%–14% of species are likely to face a very high risk of extinction at 1.5°C of global warming, reaching up to 48% at 5°C (Blunden, 2022). However, the Anthropocene era, as first described (Crutzen, 2002), has brought significant improvements in human health, safety, prosperity, and peacekeeping (Pinker, 2018).

To date, automation, as a key driver of the next technological era, holds great potential for upgrading industrial sectors and promoting inclusive and sustainable economic growth. Soft robotics possesses certain advantages over traditional robotics that make it well-suited for addressing the CPA and achieving the SDGs (Scoones et al., 2015; Matthews and Wynes, 2022; Meinshausen et al., 2022). By leveraging the flexibility, dexterity, biocompatibility/degradability, (re)-programmability of soft materials, and physical and embodied intelligence (EI), these robots can be used for monitoring and restoring complex environments, providing early warning systems for urban areas, and collecting information about changes in biodiversity and animal behavior (Laschi et al., 2016; Rossiter et al., 2017; Mazzolai and Laschi, 2020; Sitti, 2021). They can also mimic the movement and behavior of various animals and plants and can be used for pollination, seed dispersal, and soil aeration for regeneration and restoration in hard-to-reach environments (Valdes et al., 2012; Hartmann et al., 2021; Kim et al., 2021). There are already promising examples of soft robots being used to address major challenges in real-world applications that align with the SDGs and CPA (Tolley et al., 2014; Amend et al., 2016; Ng et al., 2021; Elfferich et al., 2022; Li et al., 2022).

Despite these advancements, the field of soft robotics still faces hurdles, such as producing cost-effective, environment-friendly, robust, and self-piloting robots that do not rely on hard components (The soft touch of robots, 2018). To fully harness the potential of soft robotics in promoting the SDGs and positive climate actions, careful planning and governance are needed, with the support of various stakeholders. This can result in the creation of new economic opportunities in the near term (i.e., 2030) and mid-term (i.e., 2050) and the development of new business models and regulatory frameworks that support a more sustainable future to mitigate the risks posed by climate change to the planet and its inhabitants (He et al., 2022; Skea et al., 2022).

## 2 Synergy between soft robotics and sustainable development goals

Soft robotics shows potential to significantly advance the UNs' SDGs by providing sustainable and environment-friendly automation (Grau Ruiz and O'Brolchain, 2022; Guenat et al., 2022). Soft robotics technologies are known for their high dexterity, sensitivity, and safety, as well as their ability to be customized to specific tasks and environments. We believe that soft robotics may represent a technology that can serve humanity and the planet by addressing global challenges and environmental degradation through targeted productive applications (Mengaldo et al., 2022). Figure 1 presents a thematic mapping of SDGs to soft robotic applications.

1.  *Soft robotics for urban farming*: Soft robots can improve food security and reduce poverty (SDG-2) by precision planting and harvesting crops in urban areas, promoting sustainable consumption (SDG-11), monitoring crop health, and providing safe food (SDG-3). It can also reduce the carbon footprint of the food system and farming (SDG-15) while creating jobs and promoting economic development in urban areas (SDG-8). Remarkable examples are from Cacucciolo et al. (2022) that demonstrated a robotic gripper for crop harvesting, from Junge and Hughes (2022) that proved a soft robotic digital twin for harvesting raspberries, and from Birrell et al. (2020) that showed the ability to manipulate lettuce without causing any damage to the vegetable. Although traditional robotics can also improve technological transition in agriculture (Fernandes et al., 2021), soft grippers are more versatile and robust during an interaction because they can deform, adapt, and easily control while colliding with the external world (Bonilla et al., 2014; Mazzolai et al., 2019; Navas et al., 2021).

2.  *Soft robotics for ocean preservation*: A significant role can be played in cleaning, protecting, preserving, and remedying marine biodiversity and ocean health (SDG-14), promoting sustainable use of ocean resources (SDG-12), and understanding the impact of climate change on oceans (SDG-6, 13). In the soft robotics' community, the use of soft materials and locomotion methods inspired by ocean-dwelling animals is being explored to safely gather data and study aquatic life without disrupting the delicate ecosystem (Christianson et al., 2020; Li G. et al., 2021; Aracri et al., 2021). For example, Katzschmann et al. (2018) presented a soft robotic fish species for close exploration of underwater life, equipped with cameras and remotely operated, aiming at surpassing traditional underwater vehicles.

3.  *Soft robotics for disaster response*: Soft robots can be used for search and rescue missions in the aftermath of natural disasters, promoting safety and reducing the impacts of natural disasters (SDG-11, 15), providing timely assistance to affected populations (SDG-3), re-building and repairing infrastructure, and improving transportation and communication networks (SDG-9) (Greer et al., 2018; Sadeghi et al., 2019; Luo et al., 2020). For instance, Hawkes et al. (2017) proposed a soft pneumatic robot with the ability to grow in length and dynamically adapt its shape to different terrains, including crossing small cracks, using onboard sensing of environmental stimuli.

4.  *Soft robotics for energy production*: The integration of renewable energy sources into soft robotics bodies may revolutionize energy production. Although challenging, this integration could significantly advance research for clean, versatile, and accessible energy production in nearly every location (SDG-7, 9) (Huang et al., 2019; Han and Yoon, 2021; Laschi and Mazzolai, 2021). A notable example is the study of plant-hybrid wind-energy systems inspired by living plant leaves (Meder et al.,

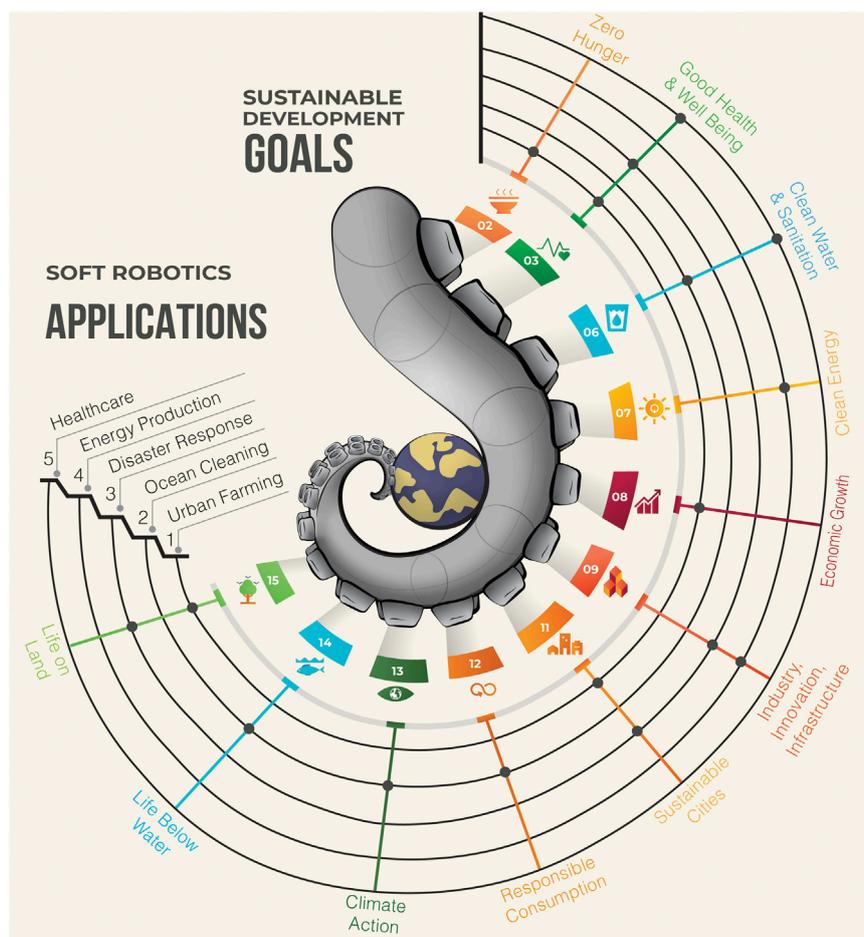

**FIGURE 1**
Illustration of an octopus arm holding the Earth with suction cups, symbolizing the boundless potential to support the United Nations' Sustainable Development Goals. The concentric lines surrounding the octopus arm represent five fields of soft robotics applications: urban farming, ocean cleaning, disaster response, energy production, and healthcare. The reader can understand the strategic mapping of SDGs to soft robotic applications. For example, soft robotics for urban farming aligns with SDGs 2, 3, 8, 11, and 15 (black dots). This correlation between SDGs and soft robot applications enhances our comprehension of the significance of future robots in promoting sustainability to enable climate transition actions.

2018). These systems convert natural mechanical stimuli, such as wind or self-touching of leaves, into electrical signals, offering potential for new energy sources that are widely distributed.

5. *Soft robotics for healthcare*: Soft robotic exosuits, wearables, and manipulators can aid in physical therapy and minimally invasive surgery (MIS), thus improving mobility capabilities, independence of individuals with disabilities, and reducing patients' recovery time (SDG-3) (di Natali et al., 2020; Poliero et al., 2020; Giordano et al., 2021b; Lin et al., 2022). In a case of MIS, Gopesh et al. (2021) developed submillimeter diameter, hydraulically actuated devices to enable active steerability of microcatheter tips. They demonstrated guidewire-free navigation, access, and coil deployment *in vivo*, offering safety and ease of use in endovascular intervention.

In this perspective, although we consider the development of functional, field-ready soft robots that support the SDGs, we would prioritize ecological sustainability in the design and fabrication processes. Future autonomous soft robots must incorporate key features such as sustainable materials and manufacturing, and environment-friendly design principles, implementing a cradle-to-cradle (C2C) approach that will help minimize the carbon footprint of the new automation era (Gao et al., 2018).

## 2.1 Biodegradable and multifunctional materials for soft robotics

Researchers envision a future where groups or individual soft robots not only perform their intended tasks but also do not harm the ecosystem throughout their entire lifecycle, thus supporting the principles of SDG-12. To achieve this goal, we believe it is essential to use biodegradable and soft multifunctional materials, whose physical and chemical properties can be manipulated to ensure intelligence while minimizing their ecological impact (Rossiter et al., 2016; Mezzenga, 2021). Families of polymers such as natural fibers, liquid crystal elastomer, and thermoplastic polyesters could be adopted as constitutive multifunctional materials for C2C robot implementation in energy generation/saving, visual responses to external/internal stimuli capabilities etc. (Shintake et al., 2017; Prévôt et al., 2018; Zaia et al., 2019; Mishra et al., 2020).

As a futuristic idea, we conceptualize a cephalopod-inspired soft robot, designed to camouflage in its surroundings to study the behavior and biology of lone aquatic animals, thus assisting in ocean conservation and management (SDG-14). Its biodegradable constitutive materials may implement intelligence such as adaptive color changing while performing continuous deformation and texturing of the body for safe interaction with living beings (Clough et al., 2020; Giordano et al., 2021a). Consistently, a system of biodegradable jellyfish-inspired robots that wirelessly and autonomously communicate may follow sea currents to gather data and develop science-based management plans aimed at restoring populations of species that are at risk due to climate change factors like acidification and nutrient depletion (Christianson et al., 2019). We believe that soft robots that are smart, biodegradable, and capable of couple sensing, actuation, computation, and communication have the potential to support SDGs 11, 13, 14, and 15 (McEvoy and Correll, 2015; Garrad et al., 2019; Kaspar et al., 2021; Meder et al., 2022). However, it is essential to shift toward a paradigm that prioritizes nature-based solutions and innovative mechanical design as central elements (Pishvar and Harne, 2020; Khaheshi and Rajabi, 2022). By extracting key parameters for multifunctionalities from natural systems that have evolved over millions of years, we can optimize the design and material of these robots to reach their maximum potential and the less consumption of raw elements (Kolle and Lee, 2018; Phillips et al., 2022).

## 2.2 Sustainable manufacturing processes

The vision of sustainable manufacturing is to create "printed future" in soft robotics, and the main advantage of additive manufacturing technologies (i.e., 3D printing) is to integrate actuators, sensors, controllers, and power systems into fully autonomous soft robots through a single, on-demand digital process (SDG-12) (Wallin et al., 2018; Helou et al., 2022; Leber et al., 2022). The "Emb3D printing" technique is an example of 3D printing to manufacture advanced soft robots (MacCurdy et al., 2016; Wehner et al., 2016). This method has been used to create untethered octobot robots that can incorporate drive systems, actuators, and soft microfluidic chip, thus showing a cost-effective solution for large-scale monitoring and swarm robotic applications. Furthermore, some robots integrate 3D printers in their body which allow them to change shape based on sensor feedback to change shape based on sensor feedback (Shah et al., 2020). For example, Sadeghi et al. (2016) developed a robot that can grow its shape through selective deposition of a biopolymer, and with several sensors at the tip, it can search for water and nutrients underground (SDG-6, 15). Recent advancements in 3D printing technology have led to the development of a rotational multi-material extruder that allows for precise control over the local orientation of architected filaments (Larson et al., 2023).

The authors envisage for future MIS applications a 5D-printed soft robot with shape-shifting capabilities that can integrate multifunctional and stimuli-responsive materials together with sensors, actuators, and artificial intelligence (AI)-driven control systems in a single manufacturing process (Pugliese and Regondi, 2022). We conceptualize the manufacturing of a biocompatible and biodegradable soft robot made of hydrogel material that can be programmed to change stiffness and shape in response to specific stimuli, such as temperature or pH. Thus, it will navigate through the body, conform to tissue, avoid obstacles, transmit data, and safely degrade. To date, few examples of existing technologies that can print multidimensional soft robots are reported by Barner-Kowollik et al. (2017), Carlotti and Mattoli (2019), and Hippler et al. (2019). Shi et al. (2019) showed a soft robot for MIS embedded with responsive materials that can be remotely controlled to perform delicate procedures, reducing invasiveness and recovery time (SDG-3, 12). Advances in sustainable manufacturing, design, and materials are driving soft robotics toward impactful solutions to achieve SDGs and aid the climate transition (Stella and Hughes, 2023).

## 2.3 Renewable energy sources for self-powered autonomous machines

The integration of renewable energy sources in soft robotics technologies holds great promise to support SDG-7 and 9. However, incorporating these sources into the soft and flexible body of the robot and developing efficient and lightweight energy storage and sourcing solutions that are able to fulfill their energy consumption are significant challenges. Research is being carried out on flexible and organic photovoltaics (PVs), 2D–3D nano-heterostructures, triboelectric micro/nano-generators (TENG), and fuel cells that generate power from renewable sources (Pomerantseva and Gogotsi, 2017; Chen et al., 2021; Hwa Jeong et al., 2021; NG et al., 2022). Great potential is on the adoption of portable, versatile, and flexible perovskite-based PVs that may find application in self-piloting drones, robots, and wearables, enabling the robot to be self-sufficient in terms of energy (Bierman et al., 2016; Horne et al., 2020; Rombach et al., 2021; Hou et al., 2022). However, to meet the goal of SDG-7, we suggest robots must be equipped with combined solutions. For example, in marine environments, a tethered solution can be used with a harvester and generator beyond the ocean surface and an inner body electro-/photo-catalytic proton exchange membrane cell to produce hydrogen for power generation (Kuang et al., 2019; Lü et al., 2022). This approach would allow for the camouflaged/swarm robotic systems to autonomously cover large kilometer squares for precise data acquisition (SDG-14).

To date, the scientific community is working toward reducing the energy consumption per robot task. TENGs have shown promise in this regard, but for consistent and efficient energy design, detailed studies are needed (Liu et al., 2021; Pang et al., 2021; Chen and Wang, 2022). The amount of energy consumption during specific tasks or activities such as movement or data collection must be identified. However, we deem that the future goal for energy autonomy in variable environments will be to replicate artificial photosynthesis by coating the skin of robots (McFadden and Al-Khalili, 2016; Chen et al., 2022). In the near term, we should focus on developing lightweight and integrated devices to meet the energy needs for autonomy through renewable energy sources, reduce energy consumption per robot task, and effectively conserve, distribute, and balance energy supplies in various environments (Phillips et al., 2018).

## 2.4 Robots with distributed sensors for monitoring and remediation

The implementation of distributed sensing systems, utilizing a group of specialized soft robots, can provide early warning for extreme events, prevent disaster, and improve forecasting models, contributing to sustainable cities and communities (SDG-11) and combating climate change (SDG-13) (Bergstrom et al., 2021; Spillman et al., 2021). We proposed biodegradable, self-governing soft machines that can survey large areas of the ocean and gather data for informed decision makers in compliance with SDG-17. The group of robots endowed with an efficient network of sensors and detectors, integrated onboard or within the constitutive materials, may report data on non-surfacing marine beings, rising temperatures, changing currents, and plastic degradation (SDG-14) (Yeh et al., 2016; Herbert-read et al., 2017; Jaffe et al., 2017; Picardi et al., 2020). Nonetheless, we envisage other roles for these machines that could survey deep sea pools to retrieve raw minerals for sustainable industry (SDG-9) (Yang S. et al., 2018; Dini et al., 2022).

In terrestrial environments, the integration of soft robots with satellites and traditional robotics can provide solutions for archaeological and heritage conservation (SDG-15). The inherent soft robotics dexterity makes them well-suited for environments of ruins, where they can safely navigate and manipulate fragile artifacts. Consistently, adopting terrestrial and aerial robotic solutions, we could gather data for seed plantation toward reforestation against desertification to capture carbon dioxide, or to feed population and local economies, and to evaluate new pathogens and vector-borne diseases (SDG-3, 8, and 13) (Wade et al., 2018; Chowdhary et al., 2019; Zhou et al., 2022). C2C self-burial systems or climbing micro-soft robots could be adopted on crops to gather targeted genetic information and provide precise monitoring and surveillance of plants and soil (SDG-8, 12, and 15) (Fiorello et al., 2021; Horton et al., 2021; Mazzolai et al., 2021). Furthermore, C2C micro-robots can be released in water sources and infrastructures to detect pollutants and/or nutrients (SDG-6). In sustainable cities, drones and climbing soft robots, in conjunction with 3D scanning, will enable point-in-time inspection and monitoring of buildings or bridges, modifying planned task in real-time (SDG-11). Also, in space, monitoring and harvesting debris can be effectively achieved using soft robotic solutions (Zhang et al., 2022). By integrating soft robotic grippers into a hard robotic frame and incorporating switchable and frequency-driven micro-patterned adhesives, it is possible to create a system that can effectively grasp, pick and place, and manipulate satellite debris on demand in the harsh conditions of outer space (Kwak et al., 2011; Dayan et al., 2021; Barnefske et al., 2022). However, the overall cost effectiveness remains an open question that can be solved by scalable fabrication techniques and repurposing (Li S. et al., 2021; Kim et al., 2023).

## 2.5 Embodied and physical intelligence for advanced soft-bodied machines

Embodied and physical intelligence plays a crucial role in developing autonomous soft robots that can effectively acquire, analyze, and utilize large amounts of data to tackle complex tasks such as navigating difficult trajectories, changing shape, and manipulating unknown and unstructured objects in various environments (Fischer et al., 2023). EI can simplify control parameters, resulting in more energy-efficient and cost-effective operations, but it requires a careful mechanical and material design to consider the interaction with the environment while performing a task (Laschi, 2022). Soft robotic systems with EI can benefit from incorporating AI-learning architecture and computational modeling to overcome the challenges posed by the complex dynamic interactions and body deformations that occur in open-world tasks (George Thuruthel et al., 2018; Thuruthel et al., 2019; Shih et al., 2020).

For instance, a group of self-sufficient and field-ready soft robots can be designed in their constitutive materials and mechanical structures specifically for exploring and restoring archaeological sites (SDG-15). These tasks require precision and safe manipulation, as well as an interaction with hard-to-reach environments and unstructured cracks. Thus, the authors envision a group of self-piloting soft robots with EI and efficient computational architecture. Each individual robot can work with the other like the "wood wide web" between trees and other organisms, to face hazardous conditions, minimize damage during retrieval, and ultimately be more cost-effective (Mazzolai et al., 2022). Although current solutions for this type of architectures are still energetically costly and in their early stages, effective solutions are expected to be developed in the future to help tackle some of humanity's grand challenges in support of some of the SDGs and climate crisis.

## 3 The soft automation era can address SDGs and positive climate actions

In our perspective, we have highlighted the great potential of soft robotics in addressing the SDGs and combating the climate crisis. We envision that functional and field-ready soft robots should incorporate key-values such as biodegradable and multifunctional materials, sustainable manufacturing processes, renewable energy sources, and physical and embodied intelligence to achieve the maximum performance that they can provide to society and the environment (Figure 2). We have discussed how soft robots may be useful in areas such as urban farming, ocean cleaning and preservation, disaster response, distributed clean energy production, and healthcare (Figure 1).

However, we acknowledge that despite the potential benefits of soft robotics in addressing the SDGs and CPA, there are limitations that must be addressed in terms of scalability, integration, and material adoption. Biodegradable soft robots should contain minimal harmful components and be economically viable. To date, cost-effective methods for producing widely accessible soft robot and strategies for recycling and repurposing existing robots to reduce costs and industrial waste are lacking. Additionally, the integration of hard robotics components may be necessary to improve the overall robustness to solve the great challenges, as shown in Figure 1. Nonetheless, the integration of onboard and lightweight energy devices is a significant challenge as achieving the efficiencies required to meet the energy consumption demands is still a struggle. Figure 2 illustrates the union pipeline section, where we can merge, for example, two of the soft robot topics to achieve a desired outcome. However, the question we would like to address is how an efficient soft robot appears when we combine all the key topics we identified, from biodegradable materials to embodied control.

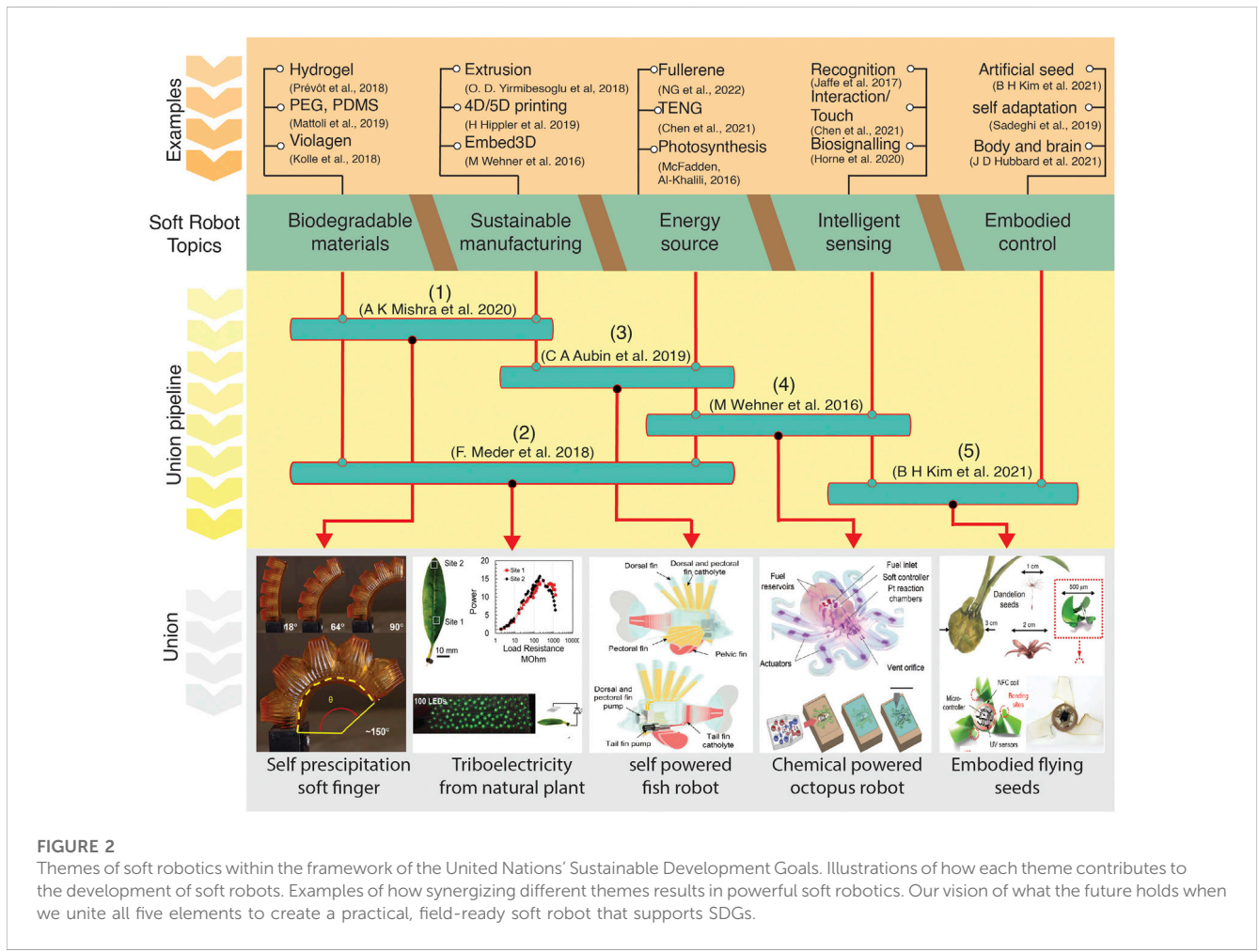

**FIGURE 2**
Themes of soft robotics within the framework of the United Nations' Sustainable Development Goals. Illustrations of how each theme contributes to the development of soft robots. Examples of how synergizing different themes results in powerful soft robotics. Our vision of what the future holds when we unite all five elements to create a practical, field-ready soft robot that supports SDGs.

To achieve such a visionary goal, we believe that interdisciplinary scientific fields must collaborate to develop environment-friendly self-piloting robots. Recent EU-funded soft robotics projects have similar goals to support the SDGs and CPA (Elfo Project, n.d.2020; HybridHeart Project, 2017; Self-Healing Soft Robotics Project, 2019; ISeed Project, 2021). Nonetheless, it is important for the widespread use of robots in industries such as agriculture to not lead to job displacement and social and economic inequality (Acemoglu and Restrepo, 2019; Kerwin, 2020). Democratization of high-tech goods and accessibility to robotic platforms, energy sources, and open-source digital technologies should counter the trend toward further concentration of economic power (Rikap and Lundvall, 2020).

In conclusion, we firmly believe that soft robotics can play a crucial role in mitigating the detrimental effects of climate change on society, the economy, and the environment, supporting some of the UN's SDGs. It is imperative for all stakeholders to collaborate to swiftly advance high-impact laboratory results in soft robotics and identify effective solutions for tackling human-induced environmental challenges and their potential future consequences (Yang G.-Z. et al., 2018; Hawkes et al., 2021). We anticipate that this call for collaboration will be met with immediate action, enabling us to craft a more sustainable and eco-friendly future through the implementation of soft robotics technology.

## Data availability statement

The original contributions presented in the study are included in the article/Supplementary Material; further inquiries can be directed to the corresponding authors.

## Author contributions

GG conceived and conceptualized the work. GG and SM elaborated the bibliographic data, wrote the paper, and equally contributed. SM conceived and illustrated the figures. BM supervised the content. All the authors critically reviewed the final work.

## Funding

GG is funded by the European Union (ERC-2021-STG, SURFACE, ID: 101039198, DOI: 10.3030/101039198). Views and opinions expressed are, however, those of the author(s) only and do not necessarily reflect those of the European Union or the European Research Council. Neither the European Union nor the granting authority can be held responsible for them.

## Acknowledgments

GG and SM want to deeply thank all the IPCC Sixth Assessment's collaborators and all the scientists and civil movements/NGO/people around the world who are spreading the voice toward solving the environmental crisis and supporting the UN's Sustainable Development Goals. Furthermore, authors would like to express their gratitude to the reviewer for his/her valuable suggestions that helped improve this paper.

## Conflict of interest

The authors declare that the research was conducted in the absence of any commercial or financial relationships that could be construed as a potential conflict of interest.

## Publisher's note

## References

Acemoglu, D., and Restrepo, P. (2019). Automation and new tasks: How technology displaces and reinstates labor. *J. Econ. Perspect* 33, 3–30. doi:10.1257/jep.33.2.3

Amend, J., Cheng, N., Fakhouri, S., and Culley, B. (2016). Soft robotics commercialization: Jamming grippers from research to product. *Soft Robot* 3, 213–222. doi:10.1089/soro.2016.0021

Aracri, S., Giorgio-Serchi, F., Suaria, G., Sayed, M. E., Nemitz, M. P., Mahon, S., et al. (2021). Soft robots for ocean exploration and offshore operations: A perspective. *Soft Robot* 8, 625–639. doi:10.1089/soro.2020.0011

Arias, P., Bellouin, N., Coppola, E., Jones, R., Krinner, G., Marotzke, J., et al. (2021). Climate change 2021: The physical science basis. Contribution of working Group14 I to the Sixth assessment report of the intergovernmental Panel on climate change. *Tech. Summ.*

Armstrong McKay, D. I., Staal, A., Abrams, J. F., Winkelmann, R., Sakschewski, B., Loriani, S., et al. (2022). Exceeding 1.5°C global warming could trigger multiple climate tipping points. *Science* 377, 377. eabn7950. doi:10.1126/science.abn7950

Barnard, P. L., Dugan, J. E., Page, H. M., Wood, N. J., Hart, J. A. F., Cayan, D. R., et al. (2021). Multiple climate change - driven tipping points for coastal systems. *Sci. Rep* 11 (1), 1–13. doi:10.1038/s41598-021-94942-7

Barnefske, L., Rundel, F., Moh, K., Hensel, R., Zhang, X., and Arzt, E. (2022). Tuning the release force of microfibrillar adhesives by geometric design. *Adv. Mater Interfaces n/a* 9, 2201232. doi:10.1002/admi.202201232

Barner-Kowollik, C., Bastmeyer, M., Blasco, E., Delaittre, G., Müller, P., Richter, B., et al. (2017). 3D laser micro- and nanoprinting: Challenges for chemistry. *Angew. Chem. Int. Ed. Engl* 56, 15828–15845. doi:10.1002/anie.201704695

Bergstrom, D. M., Wienecke, B. C., van den Hoff, J., Hughes, L., Lindenmayer, D. B., Ainsworth, T. D., et al. (2021). Combating ecosystem collapse from the tropics to the Antarctic. *Glob. Chang. Biol* 27, 1692–1703. doi:10.1111/gcb.15539

Bierman, D. M., Lenert, A., Chan, W. R., Bhatia, B., Celanović, I., Soljačić, M., et al. (2016). Enhanced photovoltaic energy conversion using thermally based spectral shaping. *Nat. Energy* 1, 16068. doi:10.1038/nenergy.2016.68

Birrell, S., Hughes, J., Cai, J. Y., and Iida, F. (2020). A field-tested robotic harvesting system for iceberg lettuce. *J. Field Robot.* 37, 225–245. doi:10.1002/rob.21888

Blunden, J., and Trewin, B. (2022). *State of the climate in 2021*. Geneva, Switzerland: WMO Library. Si–S465.

Bonilla, M., Farnioli, E., Piazza, C., Catalano, M., Grioli, G., Garabini, M., et al. (2014). "Grasping with soft hands," in 2014 IEEE-RAS International Conference on Humanoid Robots, Madrid, Spain, November 18-20, 2014, 581–587. doi:10.1109/HUMANOIDS.2014.7041421

Cacucciolo, V., Shea, H., and Carbone, G. (2022). Peeling in electroadhesion soft grippers. *Extreme Mech. Lett* 50, 101529. doi:10.1016/j.eml.2021.101529

Carlotti, M., and Mattoli, V. (2019). Functional materials for two-photon polymerization in microfabrication. *Small* 15, 1902687–1902709. doi:10.1002/smll.201902687

Chen, B., and Wang, Z. L. (2022). Toward a new era of sustainable energy: Advanced triboelectric nanogenerator for harvesting high entropy energy. *Small n/a* 18, 2107034. doi:10.1002/smll.202107034

Chen, X., Lawrence, J. M., Wey, L. T., Schertel, L., Jing, Q., Vignolini, S., et al. (2022). 3D-printed hierarchical pillar array electrodes for high-performance semi-artificial photosynthesis. *Nat. Mater* 21, 811–818. doi:10.1038/s41563-022-01205-5

Chen, Y., Mellot, G., van Luijk, D., Creton, C., and Sijbesma, R. P. (2021). Mechanochemical tools for polymer materials. *Chem. Soc. Rev* 50, 4100–4140. doi:10.1039/d0cs00940g

Chowdhary, G., Gazzola, M., Krishnan, G., Soman, C., and Lovell, S. (2019). Soft robotics as an enabling technology for agroforestry practice and research. *Sustainability* 11, 6751. doi:10.3390/su11236751

Christianson, C., Bayag, C., Li, G., Jadhav, S., Giri, A., Agba, C., et al. (2019). Jellyfish-inspired soft robot driven by fluid electrode dielectric organic robotic actuators. *Front. Robot. AI* 6, 126. doi:10.3389/frobt.2019.00126

Christianson, C., Cui, Y., Ishida, M., Bi, X., Zhu, Q., Pawlak, G., et al. (2020). Cephalopod-inspired robot capable of cyclic jet propulsion through shape change. *Bioinspir Biomim* 16, 016014. doi:10.1088/1748-3190/abbc72

Clough, J. M., van der Gucht, J., Kodger, T. E., and Sprakel, J. (2020). Cephalopod-inspired high dynamic range mechano-imaging in polymeric materials. *Adv. Funct. Mater* 30, 2002716. doi:10.1002/adfm.202002716

Crutzen, P. J. (2002). Geology of mankind. *Nature* 415, 23. doi:10.1038/415023a

Dayan, C. B., Chun, S., Krishna-Subbaiah, N., Drotlef, D.-M., Akolpoglu, M. B., and Sitti, M. (2021). 3D printing of elastomeric bioinspired complex adhesive microstructures. *Adv. Mater* 33, 2103826. doi:10.1002/adma.202103826

Di Natali, C., Sadeghi, A., Mondini, A., Bottenberg, E., Hartigan, B., de Eyto, A., et al. (2020). Pneumatic quasi-passive actuation for soft assistive lower limbs exoskeleton. *Front. Neurorobot* 31, 31. doi:10.3389/fnbot.2020.00031

Dini, A., Lattanzi, P., Ruggieri, G., and Trumpy, E. (2022). Lithium occurrence in Italy—an overview. *Minerals* 12, 945. doi:10.3390/min12080945

Elfferich, J. F., Dodou, D., and Santina, C. D. (2022). Soft robotic grippers for crop handling or harvesting: A review. *IEEE Access* 10, 75428–75443. doi:10.1109/ACCESS.2022.3190863

Elfo (2020). Elfo Project (n.d.). Available at; https://elfoproject.eu/.2020.

Fernandes, M., Scaldaferri, A., Fiameni, G., Teng, T., Gatti, M., Poni, S., et al. (2021). "Grapevine winter pruning automation: On potential pruning points detection through 2D plant modeling using grapevine segmentation," in 2021 IEEE 11th Annual International Conference on CYBER Technology in Automation, Control, and Intelligent Systems (CYBER), Jiaxing, China, July 2021, 13–18. doi:10.1109/CYBER53097.2021.9588303

Fiorello, I., Meder, F., Mondini, A., Sinibaldi, E., Filippeschi, C., Tricinci, O., et al. (2021). Plant-like hooked miniature machines for on-leaf sensing and delivery. *Commun. Mater* 2, 103–111. doi:10.1038/s43246-021-00208-0

Fischer, O., Toshimitsu, Y., Kazemipour, A., and Katzschmann, R. K. (2023). Dynamic task space control enables soft manipulators to perform real-world tasks. *Adv. Intell. Syst* 5, 2200024. doi:10.1002/aisy.202200024

Gao, M., Shih, C.-C., Pan, S.-Y., Chueh, C.-C., and Chen, W.-C. (2018). Advances and challenges of green materials for electronics and energy storage applications: From design to end-of-life recovery. *J. Mater Chem. A Mater* 6, 20546–20563. doi:10.1039/C8TA07246A

Garrad, M., Soter, G., Conn, A. T., Hauser, H., and Rossiter, J. (2019). A soft matter computer for soft robots. *Sci. Robot.* 4, eaaw6060. doi:10.1126/scirobotics.aaw6060

George Thuruthel, T., Ansari, Y., Falotico, E., and Laschi, C. (2018). Control strategies for soft robotic manipulators: A survey. *Soft Robot* 5, 149–163. doi:10.1089/soro.2017.0007

Giordano, G., Carlotti, M., and Mazzolai, B. (2021a). A perspective on cephalopods mimicry and bioinspired technologies toward proprioceptive autonomous soft robots. *Adv. Mater Technol* 6, 2100437. doi:10.1002/admt.202100437

Giordano, G., Gagliardi, M., Huan, Y., Carlotti, M., Mariani, A., Menciassi, A., et al. (2021b). Toward mechanochromic soft material-based visual feedback for electronics-free surgical effectors. *Robotics* 7, 1–16. doi:10.1002/advs.202100418

Gopesh, T., Wen, J. H., Santiago-Dieppa, D., Yan, B., Pannell, J. S., Khalessi, A., et al. (2021). Soft robotic steerable microcatheter for the endovascular treatment of cerebral disorders. *Sci. Robot.* 6, eabf0601. doi:10.1126/scirobotics.abf0601

Grau Ruiz, M. A., and O'Brolchain, F. (2022). Environmental robotics for a sustainable future in circular economies. *Nat. Mach. Intell* 4, 3–4. doi:10.1038/s42256-021-00436-6

Greer, J. D., Blumenschein, L. H., Okamura, A. M., and Hawkes, E. W. (2018). "Obstacle-aided navigation of a soft growing robot," in 2018 IEEE International Conference on Robotics and Automation (ICRA), Brisbane, Australia, May 2018, 1–8. doi:10.1109/ICRA36916.2018

Guenat, S., Purnell, P., Davies, Z. G., Nawrath, M., Stringer, L. C., Babu, G. R., et al. (2022). Meeting sustainable development goals via robotics and autonomous systems. *Nat. Commun* 1, 3559. doi:10.1038/s41467-022-31150-5

Han, M. J., and Yoon, D. K. (2021). Advances in soft materials for sustainable electronics. *Engineering* 7, 564–580. doi:10.1016/j.eng.2021.02.010

Hartmann, F., Baumgartner, M., and Kaltenbrunner, M. (2021). Becoming sustainable, the new frontier in soft robotics. *Adv. Mater* 33, 2004413. doi:10.1002/adma.202004413

Hawkes, E. W., Blumenschein, L. H., Greer, J. D., and Okamura, A. M. (2017). A soft robot that navigates its environment through growth. *Sci. Robot* 2, eaan3028. doi:10.1126/scirobotics.aan3028

Hawkes, E. W., Majidi, C., and olley, M. T. (2021). Hard questions for soft robotics. *Sci. Robot* 1–6, 1966. doi:10.1126/scirobotics.abg6049

He, C., Zhang, Y., Schneider, A., Ma, W., Kinney, P. L., and Kan, H. (2022). The inequality labor loss risk from future urban warming and adaptation strategies. *Nat. Commun* 13, 1–9. doi:10.1038/s41467-022-31145-2

Helou, C. E., Grossmann, B., Tabor, C. E., Buskohl, P. R., and Harne, R. L. (2022). Mechanical integrated circuit materials. *Nature* 608, 699–703. doi:10.1038/s41586-022-05004-5

Herbert-read, J. E., Thornton, A., Amon, D. J., Birchenough, S. N. R., Côté, I. M., Dias, M. P., et al. (2017). A global horizon scan of issues impacting marine and coastal biodiversity conservation. *Nat. Ecol. Evol* 6, 1262–1270. doi:10.1038/s41559-022-01812-0

Hippler, M., Blasco, E., Qu, J., Tanaka, M., Barner-Kowollik, C., Wegener, M., et al. (2019). Controlling the shape of 3D microstructures by temperature and light. *Nat. Commun* 10, 232. doi:10.1038/s41467-018-08175-w

Horne, J., McLoughlin, L., Bury, E., Koh, A. S., and Wujcik, E. K. (2020). Interfacial phenomena of advanced composite materials toward wearable platforms for biological and environmental monitoring sensors, armor, and soft robotics. *Adv. Mater Interfaces* 7, 1901851. doi:10.1002/admi.201901851

Horton, P., Long, S. P., Smith, P., Banwart, S. A., and Beerling, D. J. (2021). Technologies to deliver food and climate security through agriculture. *Nat. Plants* 7, 250–255. doi:10.1038/s41477-021-00877-2

Hou, Y., Yang, Y., Wang, Z., Li, Z., Zhang, X., Bethers, B., et al. (2022). Whole fabric-assisted thermoelectric devices for wearable electronics. *Adv. Sci* 9, 2103574. doi:10.1002/advs.202103574

Huang, s., Liu, Y., Zhao, Y., Ren, Z., and Guo, C. F. (2019). Flexible electronics: Stretchable electrodes and their future. *Adv. Funct. Mater* 29, 1805924. doi:10.1002/adfm.201805924

Hwa Jeong, G., Chuan Tan, Y., Tae Song, J., Lee, G.-Y., Jin Lee, H., Lim, J., et al. (2021). Synthetic multiscale design of nanostructured Ni single atom catalyst for superior CO2 electroreduction. *Chem. Eng. J* 426, 131063. doi:10.1016/j.cej.2021.131063

Hybrid Heart (2017). HybridHeart project. Available at; https://hybridheart.eu/.

Iseed (2021). ISeed project. Available at; https://iseedproject.eu/.

Jaffe, J. S., Franks, P. J. S., Roberts, P. L. D., Mirza, D., Schurgers, C., Kastner, R., et al. (2017). A swarm of autonomous miniature underwater robot drifters for exploring submesoscale ocean dynamics. *Nat. Commun* 8, 14189. doi:10.1038/ncomms14189

Junge, K., and Hughes, J. (2022). "Soft sensorized physical twin for harvesting raspberries," in 2022 IEEE 5th International Conference on Soft Robotics (RoboSoft), Edinburgh, United Kingdom, April 4-8, 2022, 601–606. doi:10.1109/RoboSoft54090.2022.9762135

Kaspar, C., Ravoo, B. J., Wiel, W. G., Wegner, S. V., and Pernice, W. H. P. (2021). The rise of intelligent matter. *Nature* 594, 345–355. doi:10.1038/s41586-021-03453-y

Katzschmann, R. K., DelPreto, J., MacCurdy, R., and Rus, D. (2018). Exploration of underwater life with an acoustically controlled soft robotic fish. *Sci. Robot* 3, eaar3449. doi:10.1126/scirobotics.aar3449

Kemp, L., Xu, C., Depledge, J., Ebi, K. L., Gibbins, G., and Kohler, T. A. (2022). Climate Endgame: Exploring catastrophic climate change scenarios. *Proc. Natl. Acad. Sci* 119, 1–9. Published. doi:10.1073/pnas.2108146119/-/DCSupplemental

Kerwin, D. (2020). International migration and work: Charting an ethical approach to the future. *J. Migr. Hum. Secur* 8, 111–133. doi:10.1177/2331502420913228

Khaheshi, A., and Rajabi, H. (2022). Mechanical intelligence (mi): A bioinspired concept for transforming engineering design. *Adv. Sci* 9, 2203783. doi:10.1002/advs.202203783

Kim, B. H., Li, K., Kim, J.-T., Park, Y., Jang, H., Wang, X., et al. (2021). Three-dimensional electronic microfliers inspired by wind-dispersed seeds. *Nature* 597, 503–510. doi:10.1038/s41586-021-03847-y

Kim, J.-K., Krishna-Subbaiah, N., Wu, Y., Ko, J., Shiva, A., and Sitti, M. (2023). Enhanced flexible mold lifetime for roll-to-roll scaled-up manufacturing of adhesive complex microstructures. *Adv. Mater* 35, 2207257. doi:10.1002/adma.202207257

Kolle, M., and Lee, S. (2018). Progress and opportunities in soft photonics and biologically inspired optics. *Adv. Mater* 30, 1702669–1702709. doi:10.1002/adma.201702669

Kuang, Y., Kenney, M. J., Meng, Y., Hung, W., Liu, Y., and Erick, J. (2019). Solar-driven, highly sustained splitting of seawater into hydrogen and oxygen fuels. *Proc. Natl. Acad. Sci. U. S. A* 116, 1–6. doi:10.1073/pnas.1900556116

Kwak, M. K., Pang, C., Jeong, H.-E., Kim, H.-N., Yoon, H., Jung, H.-S., et al. (2011). Towards the next level of bioinspired dry adhesives: New designs and applications. *Adv. Funct. Mater* 21, 3606–3616. doi:10.1002/adfm.201100982

Larson, N. M., Mueller, J., Chortos, A., Davidson, Z. S., Clarke, D. R., and Lewis, J. A. (2023). Rotational multimaterial printing of filaments with subvoxel control. *Nature* 613, 682–688. doi:10.1038/s41586-022-05490-7

Laschi, C. (2022). Embodied intelligence in soft robotics: Joys and sorrows. *IOP Conf. Ser. Mater Sci. Eng* 1261, 012002. doi:10.1088/1757-899X/1261/1/012002

Laschi, C., and Mazzolai, B. (2021). Bioinspired materials and approaches for soft robotics. *MRS Bull* 46, 345–349. doi:10.1557/s43577-021-00075-7

Laschi, C., Mazzolai, B., and Cianchetti, M. (2016). Soft robotics: Technologies and systems pushing the boundaries of robot abilities. *Sci. Robot* 1, eaah3690–12. doi:10.1126/scirobotics.aah3690

Leber, A., Dong, C., Laperrousaz, s., Banerjee, H., Abdelaziz, M. E. M. K., Bartolomei, N., et al. (2022). Highly integrated multi-material fibers for soft robotics. *Adv. Sci. n/a* 10, 2204016. doi:10.1002/advs.202204016

Li, G., Chen, X., Zhou, F., Liang, Y., Xiao, Y., Cao, X., et al. (2021a). Self-powered soft robot in the mariana trench. *Nature* 591, 66–71. doi:10.1038/s41586-020-03153-z

Li, M., Pal, A., Aghakhani, A., Pena-Francesch, A., and Sitti, M. (2022). Soft actuators for real-world applications. *Nat. Rev. Mater* 7, 235–249. doi:10.1038/s41578-021-00389-7

Li, S., Awale, S. A., Bacher, K. E., Buchner, T. J., della Santina, C., Wood, R. J., et al. (2021b). Scaling up soft robotics: A meter-scale, modular, and reconfigurable soft robotic system. *Soft Robot* 9, 324–336. doi:10.1089/soro.2020.0123

Lin, Z., Shao, Q., Liu, X.-J., and Zhao, H. (2022). An anthropomorphic musculoskeletal system with soft joint and multifilament pneumatic artificial muscles. *Adv. Intell. Syst* 4, 2200126. doi:10.1002/aisy.202200126

Liu, Y., Chen, B., Li, W., Zu, L., Tang, W., and Wang, Z. L. (2021). Bioinspired triboelectric soft robot driven by mechanical energy. *Adv. Funct. Mater* 31, 2104770. doi:10.1002/adfm.202104770

Lü, X., Deng, R., Li, X., and Wu, Y. (2022). Comprehensive performance evaluation and optimization of hybrid power robot based on proton exchange membrane fuel cell. *Int. J. Energy Res* 46, 1934–1950. doi:10.1002/er.7308

Luo, M., Wan, Z., Sun, Y., Skorina, E. H., Tao, W., Chen, F., et al. (2020). Motion planning and iterative learning control of a modular soft robotic snake. *Front. Robot. AI* 7, 599242. doi:10.3389/frobt.2020.599242

MacCurdy, R., Katzschmann, R., Kim, Y., and Rus, D. (2016). "Printable hydraulics: A method for fabricating robots by 3D co-printing solids and liquids," in 2016 IEEE International Conference on Robotics and Automation (ICRA), Stockholm, Sweden, May 2016, 3878–3885. doi:10.1109/ICRA.2016.7487576

Matthews, H. D., and Wynes, S. (2022). Current global efforts are insufficient to limit warming to 1.5° C. *Science* 376, 1404–1409. doi:10.1126/science.abo3378

Mazzolai, B., Kraus, T., Pirrone, N., Kooistra, L., de Simone, A., Cottin, A., et al. (2022). "Advancing environmental intelligence through novel approaches in soft bioinspired robotics and allied technologies: I-seed project position paper for environmental intelligence in europe," in Proceedings of the 2022 ACM Conference on Information Technology for Social Good GoodIT '22, Limassol, Cyprus, September 7-9, 2022 (New York, NY, USA: Association for Computing Machinery), 265–268. doi:10.1145/3524458.3547262

Mazzolai, B., Kraus, T., Pirrone, N., Kooistra, L., de Simone, A., Cottin, A., et al. (2021). "Towards new frontiers for distributed environmental monitoring based on an ecosystem of plant seed-like soft robots," in Proceedings of the Conference on Information Technology for Social Good, Roma Italy, September 2021, 221–224.

Mazzolai, B., and Laschi, C. (2020). A vision for future bioinspired and biohybrid robots. Sci. Robot 5, eaba6893. doi:10.1126/scirobotics.aba6893

Mazzolai, B., Mondini, A., Tramacere, F., Riccomi, G., Sadeghi, A., Giordano, G., et al. (2019). Octopus-Inspired soft arm with suction cups for enhanced grasping tasks in confined environments. Adv. Intell. Syst 1, 1900041. doi:10.1002/aisy.201900041

McEvoy, M. A., and Correll, N. (2015). Materials science. Materials that couple sensing, actuation, computation, and communication. Science 347, 1261689. doi:10.1126/science.1261689

McFadden, J., and Al-Khalili, J. (2016). Life on the edge: The coming of age of quantum biology. New York, NY, USA: Broadway Books.

Meder, F., Babu, S. P. M., and Mazzolai, B. (2022). A plant tendril-like soft robot that grasps and anchors by exploiting its material arrangement. IEEE Robot. Autom. Lett 7, 5191–5197. doi:10.1109/LRA.2022.3153713

Meder, F., Must, I., Sadeghi, A., Mondini, A., Filippeschi, C., Beccai, L., et al. (2018). Energy conversion at the cuticle of living plants. Adv. Funct. Mater 28, 1806689. doi:10.1002/adfm.201806689

Meinshausen, M., Lewis, J., Mcglade, C., Gütschow, J., Nicholls, Z., Burdon, R., et al. (2022). Realization of Paris Agreement pledges may limit warming just below 2 °C. Nature 604, 304–309. doi:10.1038/s41586-022-04553-z

Mengaldo, G., Renda, F., Brunton, S. L., Bächer, M., Calisti, M., Duriez, C., et al. (2022). A concise guide to modelling the physics of embodied intelligence in soft robotics. Nat. Rev. Phys 4, 595–610. doi:10.1038/s42254-022-00481-z

Mezzenga, R. (2021). Grand challenges in soft matter. Front. Soft Matter 1. doi:10.3389/frsfm.2021.811842

Mishra, A. K., Wallin, T. J., Pan, W., Xu, P., Wang, K., Giannelis, E. P., et al. (2020). Autonomic perspiration in 3D-printed hydrogel actuators. Sci. Robot 5, eaaz3918. doi:10.1126/scirobotics.aaz3918

Navas, E., Fernández, R., Sepúlveda, D., Armada, M., and Gonzalez-de-Santos, P. (2021). Soft grippers for automatic crop harvesting: A review. Sensors 21, 2689. doi:10.3390/s21082689

Ng, C. S. X., Tan, M. W. M., Xu, C., Yang, Z., Lee, P. S., and Lum, G. Z. (2021). Locomotion of miniature soft robots. Adv. Mater 33, 2003558. doi:10.1002/adma.202003558

Ng, L. W. T., Lee, S. W., Chang, D. W., Hodgkiss, J. M., and Vak, D. (2022). Organic photovoltaics' new renaissance: Advances toward roll-to-roll manufacturing of non-fullerene acceptor organic photovoltaics. Adv. Mater Technol. n/a 7, 2101556. doi:10.1002/admt.202101556

Pang, Y., Cao, Y., Derakhshani, M., Fang, Y., Wang, Z. L., and Cao, C. (2021). Hybrid energy-harvesting systems based on triboelectric nanogenerators. Matter 4, 116–143. doi:10.1016/j.matt.2020.10.018

Phillips, B. T., Becker, K. P., Kurumaya, S., Galloway, K. C., Whittredge, G., Vogt, D. M., et al. (2018). A dexterous, glove-based teleoperable low-power soft robotic arm for delicate deep-sea biological exploration. Sci. Rep 8, 14779. doi:10.1038/s41598-018-33138-y

Phillips, J. W., Prominski, A., and Tian, B. (2022). Recent advances in materials and applications for bioelectronic and biorobotic systems. View 3, 1–15. doi:10.1002/VIW.20200157

Picardi, G., Chellapurath, M., Iacoponi, S., Stefanni, S., Laschi, C., and Calisti, M. (2020). Bioinspired underwater legged robot for seabed exploration with low environmental disturbance. Sci. Robot 5, eaaz1012. doi:10.1126/scirobotics.aaz1012

Pinker, S. (2018). Enlightenment now: The case for reason, science, humanism, and progress. Penguin, United Kingdom: Penguin Books Limited.

Pishvar, M., and Harne, R. L. (2020). Foundations for soft, smart matter by active mechanical metamaterials. Adv. Sci 7, 2001384. doi:10.1002/advs.202001384

Poliero, T., Lazzaroni, M., Toxiri, S., di Natali, C., Caldwell, D. G., and Ortiz, J. (2020). Applicability of an active back-support exoskeleton to carrying activities. Front. Robot. AI 7, 579963. doi:10.3389/frobt.2020.579963

Pomerantseva, E., and Gogotsi, Y. (2017). Two-dimensional heterostructures for energy storage. Nat. Energy 2, 17089. doi:10.1038/nenergy.2017.89

Pörtner, H.-O., Roberts, D. C., Adams, H., Adler, C., Aldunce, P., Ali, E., et al. (2022). Geneva, Switzerland: IPCC. Sixth Assessment Report. Climate change 2022: Impacts, adaptation and vulnerability

Prévôt, M. E., Ustunel, S., and Hegmann, E. (2018). Liquid crystal elastomers—A path to biocompatible and biodegradable 3D-LCE scaffolds for tissue regeneration. Materials 11, 377. doi:10.3390/ma11030377

Pugliese, R., and Regondi, S. (2022). Artificial intelligence-empowered 3D and 4D printing technologies toward smarter biomedical materials and approaches. Polym. (Basel) 14, 2794. doi:10.3390/polym14142794

Rikap, C., and Lundvall, B.-Å. (2020). Big tech, knowledge predation and the implications for development. Innovation Dev 12, 389–416. doi:10.1080/2157930x.2020.1855825

Rombach, F. M., Haque, S. A., and Macdonald, T. J. (2021). Lessons learned from spiro-OMeTAD and PTAA in perovskite solar cells. Energy Environ. Sci 14, 5161–5190. doi:10.1039/D1EE02095A

Rossiter, J., Winfield, J., and Ieropoulos, I. (2017). "Eating, drinking, living, dying and decaying soft robots," in Soft robotics: Trends, applications and challenges (Singapore: Springer), 95–101.

Rossiter, J., Winfield, J., and Ieropoulos, I. (2016). "Here today, gone tomorrow: Biodegradable soft robots," in Electroactive Polymer Actuators and Devices (EAPAD) 2016 (International Society for Optics and Photonics), Las Vegas, United States, 21-24 March 2016, 97981S.

Sadeghi, A., del Dottore, E., Mondini, A., and Mazzolai, B. (2019). Passive morphological adaptation for obstacle avoidance in a self-growing robot produced by additive manufacturing. Soft Robot 7, 85–94. doi:10.1089/soro.2019.0025

Sadeghi, A., Mondini, A., del Dottore, E., Mattoli, V., Beccai, L., Taccola, S., et al. (2016). A plant-inspired robot with soft differential bending capabilities. Bioinspir Biomim 12, 015001. doi:10.1088/1748-3190/12/1/015001

Scoones, I., Leach, M., and Newell, P. (2015). The politics of green transformations. Oxfordshire, United Kingdom: Taylor & Francis.

Self-Healing (2019). A Horizon 2020 FET Open Project. Available at: https://www.sherofet.eu/.

Shah, D., Yang, B., Kriegman, S., Levin, M., Bongard, J., and Kramer-Bottiglio, R. (2020). Shape changing robots: Bioinspiration, simulation, and physical realization. Adv. Mater. n/a 33, 2002882. doi:10.1002/adma.202002882

Shi, Q., Liu, H., Tang, D., Li, Y., Li, X., and Xu, F. (2019). Bioactuators based on stimulus-responsive hydrogels and their emerging biomedical applications. NPG Asia Mater 11, 64. doi:10.1038/s41427-019-0165-3

Shih, B., Shah, D., Li, J., Thuruthel, T. G., Park, Y.-L., Iida, F., et al. (2020). Electronic skins and machine learning for intelligent soft robots. Sci. Robot 5, eaaz9239. doi:10.1126/scirobotics.aaz9239

Shintake, J., Sonar, H., Piskarev, E., Paik, J., and Floreano, D. (2017). "Soft pneumatic gelatin actuator for edible robotics," in 2017 IEEE/RSJ International Conference on Intelligent Robots and Systems (IROS), Vancouver, Canada, September 24-1, 2017, 6221–6226. doi:10.1109/IROS.2017.8206525

Sitti, M. (2021). Physical intelligence as a new paradigm. Extreme Mech. Lett 46, 101340. doi:10.1016/j.eml.2021.101340

Skea, J., Shukla, P., and Kılkış, Ş. (2022). Climate change 2022: Mitigation of climate change. Geneva, Switzerland: IPCC.

Spillman, C. M., Smith, G. A., Hobday, A. J., and Hartog, J. R. (2021). Onset and decline rates of marine heatwaves: Global trends, seasonal forecasts and marine management. Front. Clim 3, 182. doi:10.3389/fclim.2021.801217

Stella, F., and Hughes, J. (2023). The science of soft robot design: A review of motivations, methods and enabling technologies. Front Robot AI 9 Available at: https://www.frontiersin.org/articles/10.3389/frobt.2022.1059026.

The soft touch of robots (2018). Nat. Rev. Mater 3, 71. doi:10.1038/s41578-018-0017-8

Thuruthel, T. G., Shih, B., Laschi, C., and Tolley, M. T. (2019). Soft robot perception using embedded soft sensors and recurrent neural networks. Sci. Robot 4, eaav1488. doi:10.1126/scirobotics.aav1488

Tolley, M. T., Shepherd, R. F., Mosadegh, B., Galloway, K. C., Wehner, M., Karpelson, M., et al. (2014). A resilient, untethered soft robot. Soft Robot 1, 213–223. doi:10.1089/soro.2014.0008

United Nations Environment Programme (2016). The paris agreement. Available at: https://unfccc.int/sites/default/files/english_paris_agreement.pdf.

United Nations' Sustainable Development Goals (2023). United Nations' sustainable development goals. Available at: https://sdgs.un.org/goals.

Valdes, S., Urza, I., Pounds, P., and Singh, S. (2012). "Samara: Low-cost deployment for environmental sensing using passive autorotation," in Robotics: Science and Systems Workshop on Robotics for Environmental Monitoring (Citeseer), Sydney, Australia, July 2012.

Wade, T. I., Ndiaye, O., Mauclaire, M., Mbaye, B., Sagna, M., Guissé, A., et al. (2018). Biodiversity field trials to inform reforestation and natural resource management strategies along the African Great Green Wall in Senegal. New (Dordr) 49, 341–362. doi:10.1007/s11056-017-9623-3

Wallin, T. J., Pikul, J., and Shepherd, R. F. (2018). 3D printing of soft robotic systems. Nat. Rev. Mater 3, 84–100. doi:10.1038/s41578-018-0002-2

Wehner, M., Truby, R. L., Fitzgerald, D. J., Mosadegh, B., Whitesides, G. M., Lewis, J. A., et al. (2016). An integrated design and fabrication strategy for entirely soft, autonomous robots. *Nature* 536, 451–455. doi:10.1038/nature19100

Williams, A. I. L., Stier, P., Dagan, G., and Watson-parris, D. (2022). Strong control of effective radiative forcing by the spatial pattern of absorbing aerosol. *Nat. Clim. Chang* 12, 735–742. doi:10.1038/s41558-022-01415-4

Yang, G.-Z., Bellingham, J., Dupont, P. E., Fischer, P., Floridi, L., Full, R., et al. (2018a). The grand challenges of *Science Robotics*. *Sci. Robot* 3, eaar7650. doi:10.1126/scirobotics.aar7650

Yang, S., Zhang, F., Ding, H., He, P., and Zhou, H. (2018b). Lithium metal extraction from seawater. *Joule* 2, 1648–1651. doi:10.1016/j.joule.2018.07.006

Yeh, X., Brantner, G., Stuart, H., Wang, S., Cutkosky, M., Edsinger, A., et al. (2016). Ocean one: A robotic avatar for oceanic discovery. *IEEE Robot. Autom. Mag* 23, 20–29. doi:10.1109/MRA.2016.2613281

Zaia, E. W., Gordon, M. P., Yuan, P., and Urban, J. J. (2019). Progress and perspective: Soft thermoelectric materials for wearable and internet-of-things applications. *Adv. Electron Mater* 5, 1800823. doi:10.1002/aelm.201800823

Zhang, Y., Li, P., Quan, J., Li, L., Zhang, G., and Zhou, D. (2022). Advanced intelligent system, 1–25. doi:10.1002/aisy.202200071Progress, challenges, and prospects of soft robotics for space applications

Zhou, X., Wen, X., Wang, Z., Gao, Y., Li, H., Wang, Q., et al. (2022). Swarm of micro flying robots in the wild. *Sci. Robot* 7, eabm5954. doi:10.1126/scirobotics.abm5954